\documentclass{article}

\usepackage{times}
\usepackage{graphicx} 
\usepackage{subfigure} 

\usepackage{natbib}

\usepackage{algorithm}
\usepackage{algorithmic}

\usepackage{hyperref}

\usepackage[accepted]{icml2014}

\usepackage{amsmath,amsfonts,amsthm,amssymb}
\usepackage{symbols}

\newcommand{\MF}{\mathrm{MF}}

\icmltitlerunning{Mean Field Networks}

\begin{document} 

\twocolumn[
\icmltitle{Mean Field Networks}

\icmlauthor{Yujia Li${}^1$}{yujiali@cs.toronto.edu}
\icmlauthor{Richard Zemel${}^{1,2}$}{zemel@cs.toronto.edu}
\icmladdress{${}^1$Department of Computer Science, University of Toronto,
    Toronto, ON, Canada}
\vspace{-9pt}
\icmladdress{${}^2$Canadian Institute for Advanced Research, Toronto, ON,
Canada}

\icmlkeywords{Mean Field, Graphical Models, Neural Networks, Back Propagation,
Inference, Learning}

\vskip 0.3in
]

\begin{abstract} 
    The mean field algorithm is a widely used approximate inference algorithm
    for graphical models whose exact inference is intractable. In each
    iteration of mean field, the approximate marginals for each variable are 
    updated by getting information from the neighbors. This process can be
    equivalently converted into a feed-forward network, with each layer
    representing one iteration of mean field and with tied weights on all
    layers. This conversion enables a few natural extensions, \eg untying the
    weights in the network. In this paper, we study
    these mean field networks (MFNs), and use them as inference tools as well as
    discriminative models. Preliminary experiment results show that MFNs can
    learn to do inference very efficiently and perform significantly better
    than mean field as discriminative models.
\end{abstract} 

\section{Mean Field Networks}

In this paper, we consider pairwise MRFs defined for
random vector $\xv$ on
graph $G=(\mcV, \mcE)$ with vertex set $\mcV$ and edge set $\mcE$ of the following form,
\begin{equation}
    p(\xv) = \frac{1}{Z} \exp(E(\xv; \theta)),
\end{equation}
where the energy function $E(\xv; \theta)$ is a sum of unary ($f_s$) and pairwise
($f_{st}$) potentials
\begin{equation}
    E(\xv; \theta) = \sum_{s\in \mcV} f_s(x_s;\theta) + \sum_{(s,t)\in \mcE}
    f_{st}(x_s, x_t; \theta)
\end{equation}
$\theta$ is a set of parameters in $E$ and $Z=\sum_\xv \exp(E(\xv; \theta))$
is a normalizing constant. We assume for all
$s\in \mcV$, $x_s$ takes values from a discrete set $\mcX$, with $|\mcX|=K$.
Note that $p(\xv)$ can be a posterior distribution $p(\xv|\yv)$ (a CRF) conditioned on some
input $\yv$, and the energy function can be a function of $\yv$ with
parameter $\theta$. We do not make this dependency explicit for simplicity of
notation, but all discussions in this paper apply to conditional
distributions just as well and most of our applications are for conditional
models. Pairwise MRFs are widely used in, for example, image segmentation,
denoising, optical flow estimation, etc. Inference in such models is hard in general.

The mean field algorithm is a widely used approximate inference algorithm. The
algorithm finds the best factorial distribution $q(\xv)=\prod_{s\in\mcV}
q_s(x_s)$ that minimizes the KL-divergence with the original distribution
$p(\xv)$. The standard strategy to minimize this KL-divergence is coordinate
descent. When fixing all variables except $x_s$, the optimal distribution
$q^*_s(x_s)$ has a closed form solution 
\par\nobreak
{\small
    \setlength{\abovedisplayskip}{-5pt}
    \setlength{\belowdisplayskip}{-\abovedisplayskip}
\begin{equation}\label{eqn:mfupdate}
    q^*_s(x_s) = \frac{1}{Z_s}\exp\left(f_s(x_s;\theta) + \sum_{t\in \mcN(s)} \sum_{x_t}
    q_t(x_t) f_{st}(x_s, x_t;\theta)\right)
\end{equation}}%
where $\mcN(s)$ represents the neighborhood of vertex $s$ and $Z_s$ is a
normalizing constant. In each iteration of mean field, the $q$ distributions
for all variables are updated in turn and the algorithm is executed until some
convergence criterion is met.

\begin{figure}[tb]
    \centering
    \includegraphics[width=0.42\columnwidth]{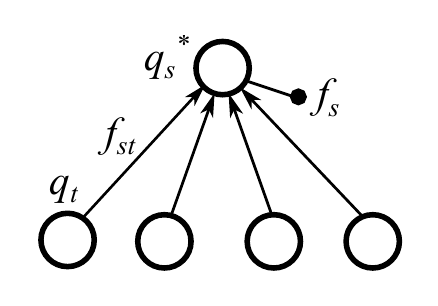}
    \vspace{-10pt}
    \caption{Illustration of one unit in Mean Field Networks.}
    \label{fig:node}
\end{figure}

We observe that \eqref{eqn:mfupdate} can be interpreted as a feed-forward
operation similar to those used in neural networks. More specifically, $q^*_s$
corresponds to the output of a node and $q_t$'s are the outputs of the layer
below, $f_s$ are biases and $f_{st}$ are weights, and the nonlinearity for
this node is a softmax function. \figref{fig:node} illustrates this
correspondence. Note that unlike ordinary neural networks, the $q$ nodes and
biases are all vectors, and the connection weights are matrices.

Based on this observation, we can map a $M$-iteration mean field algorithm
to a $M$-layer feed-forward network. Each iteration corresponds to the
forward mapping from one layer to the next, and all layers share the
same set of weights and biases given by the underlying graphical model. The
bottom layer contains the initial distributions. We
call this type of network a Mean Field Network (MFN).

\begin{figure}[tb]
    \centering
    \begin{tabular}{cc}
        \includegraphics[width=0.46\columnwidth]{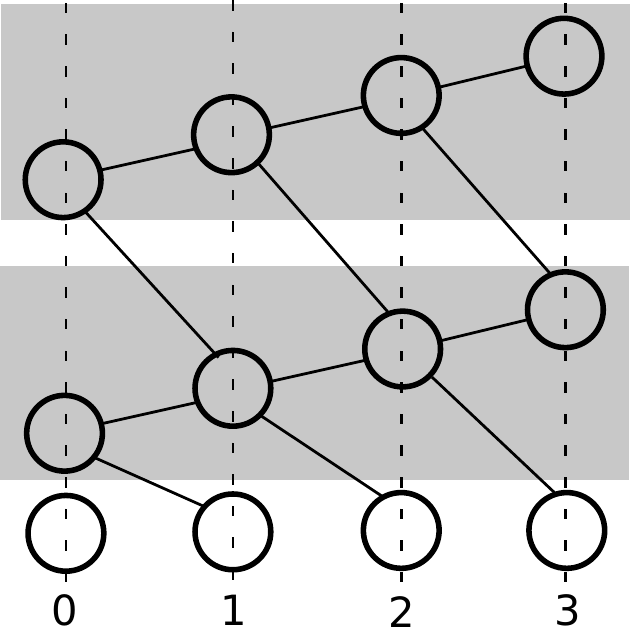} &
        \includegraphics[width=0.46\columnwidth]{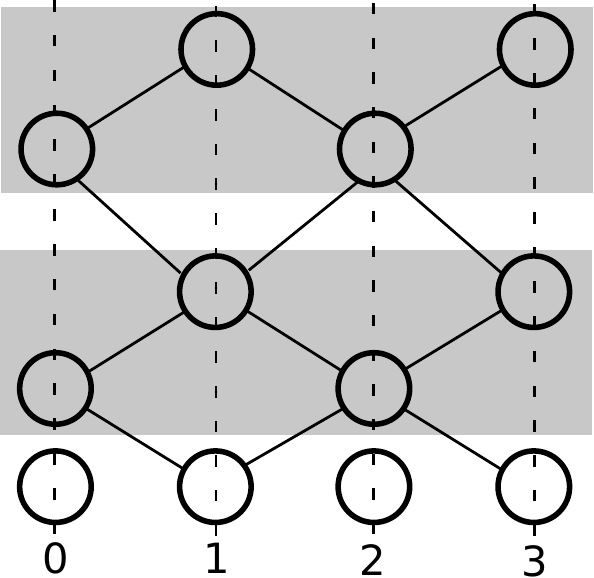} \\
        (a) & (b)
    \end{tabular}
    \vspace{-10pt}
    \caption{2-layer MFNs for a chain model
    \circletext{0}-\circletext{1}-\circletext{2}-\circletext{3} with (a)
    sequential update schedule, (b) block parallel update schedule. The
    arrows, weights and biases are dropped. The grey plates indicate
    layers. The height of a node indicates its order in the updates.}
    \label{fig:mfnchain}
\end{figure}

\figref{fig:mfnchain} shows 2-layer MFNs for a chain of 4 variables with
different update schedule in mean field. Though it is possible to do exact
inference for chain models, we use them here just for illustration. Note that the
update schedule determines the structure of the corresponding MFN.
\figref{fig:mfnchain}(a) corresponds to a sequential update schedule and
\figref{fig:mfnchain}(b) corresponds to a block parallel update schedule.

From the feed-forward network point of view, MFNs are just a special type of
feed-forward networks, with a few important restrictions on the network:
\begin{itemize}
    \vspace{-5pt}
    \item The weights and biases, or equivalently the parameter $\theta$'s,
        on all layers are tied and equal to the $\theta$ in the underlying
        pairwise MRF.
    \vspace{-5pt}
    \item The network structure is the same on all layers and follows the
        structure of the pairwise MRF.
    \vspace{-5pt}
\end{itemize}
These two restrictions make $M$-layer MFNs exactly equivalent to $M$ iterations
of the mean field algorithm. But from the feed-forward network viewpoint,
nothing stops us from relaxing the restrictions, as long as we keep the number
of outputs at the top layer constant.

By relaxing the restrictions, we lose the equivalence to mean field, but if
all we care about is the quality of the input-to-output mapping, measured by some loss
function like KL-divergence, then this relaxation can be beneficial. We
discuss a few relaxations here that aim to improve $M$-layer MFNs with fixed
$M$ as an inference tool for a pairwise MRF with fixed $\theta$:

(1) Untying $\theta$'s in MFNs from the $\theta$ in the original pairwise MRF.
If we consider $M$-layer MFNs with fixed $M$, then this relaxation can be
beneficial as the mean field algorithm is designed to run until convergence,
but not for a specific $M$. Therefore chosing some $\theta'\neq \theta$ may
lead to better KL-divergence in $M$ steps when $M$ is small. This can save
time as the same quality outputs are obtained with less steps. As $M$ grows,
we expect the optimal $\theta'$ to approach $\theta$.

(2) Untying $\theta$'s on all layers, \ie allow different $\theta$'s on
different layers. This will create a strictly more powerful model with many
more parameters. The $\theta$'s on different layers can therefore focus on
different things; for example, the lower layers can focus on
getting to a good area quickly and the higher layers can focus on
converging to an optimum fast.

(3) Untying the network structure from the underlying graphical model. If we
remove connections from the MFNs, the forward pass in the network can
be faster. If we add connections, we create a strictly more powerful model.
Information flows faster on networks with long range
connections, which is usually helpful. We can further untie the network structure on all layers, i.e. allow different layers
to have different connection structures. This creates a strictly
more flexible model.

As an example, we consider relaxation (1) for a trained pairwise CRF with
parameter $\theta$. As the model is conditioned on input data, the potentials
will be different for each data case, but the same parameter $\theta$ is used
to compute the potentials. The aim here is to use a different set of parameters $\theta'$
in MFNs to speed up inference for the CRF with parameter $\theta$ at test time, or equivalently to obtain better outputs
within a fixed inference budget. To get $\theta'$, we compute the potentials
for all data cases first using $\theta$. Then the distributions defined by
these potentials are used as targets, and we train our MFN to minimize the
KL-divergence between the outputs and the targets. Using KL-divergence as the
loss function, this training can be done by following the gradients of
$\theta'$, which can be computed by the standard back-propagation algorithm
developed for feed-forward networks. To be more specific, the KL-divergence
loss is defined as
\par\nobreak
{\small
    \setlength{\abovedisplayskip}{-5pt}
    \setlength{\belowdisplayskip}{-\abovedisplayskip}
    \setlength{\abovedisplayshortskip}{0pt}
    \setlength{\belowdisplayshortskip}{3pt}
\begin{align}
    & \KL(q^M||p) = \sum_{s\in \mcV} \sum_{x_s\in \mcX}q^M_s(x_s)\log
    q^M_s(x_s)
    - \sum_{s\in \mcV} \sum_{x_s\in\mcX} q^M_s(x_s) f_s(x_s) \nonumber\\
    &-
    \sum_{(s,t)\in\mcE}\sum_{x_s,x_t\in\mcX}q^M_s(x_s)q^M_t(x_t) f_{st}(x_s,
    x_t) + C 
\end{align}}%
where $q^M$ is the $M$th layer output and $C$ is a constant representing terms
that do not depend on $q^M$. The gradient of the loss with respect to
$q^M_s(x_s)$ can be computed as 
\par\nobreak
{\small
    \setlength{\abovedisplayskip}{-5pt}
    \setlength{\belowdisplayskip}{-\abovedisplayskip}
\begin{equation}
    \pdiff{\KL}{q^M_s(x_s)} = \log q^M_s(x_s) + 1 - f_s(x_s) - \sum_{t\in
    \mcN(s)} \sum_{x_t\in\mcX} q^M_t(x_t) f_{st}(x_s, x_t)
\end{equation}}%
The gradient with respect to $\theta'$ follows from the chain rule, as $q^M$ is a
function of $\theta'$. 

At test time, $\theta'$ instead of $\theta$ is used to compute the outputs, which
is expected to get to the same results as using mean field in fewer steps.

The discussions above focus on making MFNs better tools for inference. We can,
however, take a step even further, to abandon the underlying pairwise MRF and
use MFNs directly as discriminative models. For this setting, MFNs correspond
to conditional distributions of form $q_{\theta'}(\xv|\yv)$ where $\yv$ is
some input and $\theta'$ is the parameters. The $q$ distribution is factorial, and defined
by a forward pass of the network. The weights and biases on all layers as well as the
initial distribution at the bottom layer can depend on $\yv$ via functions
with parameters $\theta'$. These discriminative MFNs can be learned using a
training set of $(\hat{\xv}, \hat{\yv})$ pairs to minimize some loss function.
An example is the element-wise hinge loss, which is better defined on inputs
to the output layers $a^*_s(x_s)=f_s(x_s) +
\sum_{t\in\mcN(s)}\sum_{x_t}q_t(x_t)f_{st}(x_s, x_t)$, \ie the exponent part
in \eqref{eqn:mfupdate}
\par\nobreak
{\small
    \setlength{\abovedisplayskip}{-5pt}
    \setlength{\belowdisplayskip}{-\abovedisplayskip}
\begin{equation}
    L(a^M, \hat{\xv}) = \sum_{s\in\mcV} \left[\max_k \left\{a^M_s(k) + \Delta(k,
    \hat{y}_s)\right\} - a^M_s(\hat{y}_s)\right]
\end{equation}}%
where $\Delta$ is the task loss function. An example is 
$\Delta(k,\hat{y}_s)=c\Iv[k\neq \hat{y}_s]$, where $c$ is the loss for
mislabeling and $\Iv[.]$ is the indicator function. The gradient of this loss
with respect to $a^M$ has a very simple form
\begin{equation}
    \pdiff{L}{a^M_s(k)} = \Iv[k=k^*] - \Iv[k=\hat{y}_s]
\end{equation}
where $k^*=\argmax_k \left\{a^M_s(k) + \Delta(k, \hat{y}_s)\right\}$.
The gradient of $\theta'$ can then be computed using back-propagation.

Compared to the standard paradigm that uses intractable inference during
learning, these discriminative MFNs are trained with fixed inference budget
($M$ steps/layers) in
mind, and therefore can be expected to work better when we only run the
inference for a fixed number of steps. The discriminative formulation also
enables the use of a variety of different loss functions more suitable for
discriminative tasks like the hinge loss defined above, which is usually not
straight-forward to be integrated into the standard paradigm.
Many relaxations described before can be used here to make the discriminative model more powerful, for
example untying weights on different layers.

\section{Related Works}
Previous work by Justin Domke \cite{domke2011parameter, domke2013learning}
and Stoyanov \etal \cite{stoyanov2011empirical}
are the most related to ours. In \cite{domke2011parameter, domke2013learning}, the author described the idea of
truncating message-passing at learning and test time to a fixed number of
steps, and back-propagating through the truncated inference procedure to update
parameters of the underlying graphical model. In \cite{stoyanov2011empirical}
the authors proposed to train graphical models in a discriminative fashion to
directly minimize empirical risk, and used back-propagation to optimize the
graphical model parameters.

Compared to their approaches, our MFN model is one step further. 
The MFNs have a more explicit connection to feed-forward neural networks, which makes
it clear to see where the restrictions of the model are, and also more 
straight-forward to derive gradients for back-propagation. MFNs enables some 
natural relaxations of the restrictions like weight sharing, which leads to 
faster and better inference as well as more powerful prediction models.
When restricting our MFNs to have the same weights and biases on all layers
and tied to the underlying graphical model, we can recover the method in
\cite{domke2011parameter, domke2013learning} for mean field.

Another work by \cite{jain2007supervised} briefly draws a connection between mean
field inference of a specific binary MRF with neural networks, but did
not explore further variations.

A few papers have discussed the compatibility between learning and approximate
inference algorithms theoretically. \cite{wainwright2006estimating} shows that
inconsistent learning may be beneficicial when approximate inference is used
at test time, as long as the learning and test time inference are properly
aligned. \cite{kulesza2007structured} on the other hand shows that even when
using the same approximate inference algorithm at training and test time can
have problematic results when the learning algorithm is not compatible with
inference. MFNs do not have this problem, as training follows the exact
gradient of the loss function.

On the neural networks side, people have tried to use a neural network to
approximate intractable posterior distributions for a long time, especially for learning
sigmoid belief networks, see for example \cite{dayan1995helmholtz} and recent
paper \cite{mnih2014neural} and citations therein. As far as we know, no
previous work on the neural network side have discussed the connection with
mean field or belief propagation type methods used for variational inference
in graphical models.

A recent paper \cite{korattikara2014austerity} develops approximate MCMC
methods with limited inference budget, which shares the spirit of our work.

\section{Preliminary Experiment Results}

\begin{figure}[tb]
    \centering
    \includegraphics[width=0.8\columnwidth]{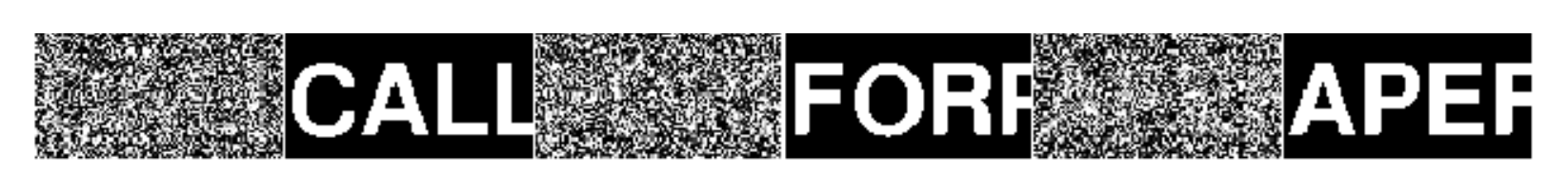}
    \vspace{-5pt}
    \caption{Three pairs of example images, in each pair: left image is the
    noisy input image, right image is the ground truth label.}
    \vspace{-10pt}
    \label{fig:data}
\end{figure}

We demonstrate the performance of MFNs on an image denoising task. We
generated a synthetic dataset of 50$\times$100 images. Each image has a black
background (intensity 0) and some random white (intensity 1) English letters as foreground. Then 
flipping noise (pixel intensity fliped from 0 to 1 or
1 to 0) and Gaussian noise are added to each pixel. The task is to recover the
clean text images from the noisy images, more specifically, to label each pixel
into one of two classes: foreground or background. In this way it is also a
binary segmentation problem. We generated training and test sets, each
containing 50 images. A few example images and corresponding labels are
shown in \figref{fig:data}.

The baseline model we consider in the experiments is a pairwise CRF. The model
defines a posterior distribution of output label $\xv$ given input image
$\yv$. For each pixel $s$ the label $x_s\in\{0, 1\}$. The conditional unary potentials are
defined using a linear model $f_s(x_s; \yv) = x_s\wv^\top\phi(\yv, s)$, where
$\phi(\yv, s)$ extracts a 5$\times$5 window around pixel $s$ and padded with a
constant 1 to form a 26-dimensional feature vector, $\wv$ is the parameter
vector for unary potentials. The pairwise potentials are defined as Potts potentials, $f_{st}(x_s, x_t;\yv)
= p_{st} \Iv[x_s=x_t]$, where $p_{st}$ is the penalty for pixel $s$ and $t$ to
take different labels. We use one single penalty $p_h$ for all horizontal
edges and another $p_v$ for all vertical edges. In total, the baseline model specified by $\theta = (\wv, p_h, p_v)$ has 28 parameters.

For all inference procedures in the experiments for both mean field and MFNs,
the distributions are initialized by taking softmax of unary potentials.

We learn $\theta$ for the baseline model by gradient ascent to maximize the
conditional log likelihood of training data. To compute the gradients, the
posterior expectations are approximated using marginals obtained by running
mean field for 30 steps (abbrev.~MF-30). $\theta$ is initialized as an all 1 vector, except
that the weight for constant feature in unary model is set to
$-5\times5/2=-12.5$. We denote this initial parameter setting as $\theta_0$,
and the parameters after training as $\theta_\MF$. With MF-30, $\theta_0$ achieves an accuracy of 0.7957 on
test set, after training, the accuracy improves to 0.8109.

\subsection{MFN for Inference}

In the first experiment, we learn MFNs to do inference for the
CRF model with parameter $\theta_\MF$. We train $M$-layer MFNs (MFN-$M$) with
fully untied weights on all layers to minimize the
KL-divergence loss for $M=1,3,10,30$. The MFN parameters on all layers are initialized to be the same as $\theta_\MF$.

As baselines, the average KL-divergence on test set using MF-1, MF-3, MF-10
and MF-30 are $-12779.05$, $-12881.50$, $-12904.43$, $-12908.54$. Note
these numbers are the KL-divergence without the constant corresponding to
log-partition function, which we cannot compute. The corresponding
KL-divergence on test set for MFN-1, MFN-3, MFN-10, MFN-30 are $-12837.87$,
$-12893.52$, $-12908.80$, $-12909.34$. We can see that MFNs improve
performance more significantly when $M$ is small, and MFN-10 is even better than MF-30, while
MF-30 runs the inference for 20 more iterations than MFN-10.

%

\subsection{MFN as Discriminative Model}

In the second experiment, we train MFNs as discriminative models for the
denoising task directly. We start with a three-layer MFN
with tied weights (MFN-3-t). The MFN parameters are initialized to be the same as
$\theta_\MF$. As baselines, MF-3 with $\theta_\MF$ achieves an accuracy of
0.8065 on test set, and MF-30 with $\theta_0$ and $\theta_\MF$ achieves
accuracy 0.7957 and 0.8109 respectively as mentioned before.

We learn MFN-3-t to minimize the element-wise hinge loss with learning rate
0.0005 and momentum 0.5. After 50 gradient steps, the test accuracy improves
and converges to around 0.8134, which beats all the mean field
baselines and is even better than MF-30 with $\theta_\MF$.

Then we untie the weights of the three-layer MFN (denoted MFN-3) and continue
training with larger learning rate 0.002 and momentum 0.9 for another 200
steps. The test accuracy improves further to around 0.8151. During learning,
we observe that the gradients for the three layers are usually quite different:
the first and third layer gradients are usually much larger than the second
layer gradients. This may cause a problem for MFN-3-t, which is essentially
using the same gradient (sum of gradients on three layers) for all three layers.

As a comparison, we tried to continue training MFN-3-t without untying
the weights using learning rate 0.002 and momentum 0.9. The test accuracy
improves to around 0.8145 but oscillated a lot and eventually diverged. We've
tried a few smaller learning rate and momentum settings but can not get the
same level of performance as MFN-3 within 200 steps.

\section{Discussion and Ongoing Work}

In this paper we proposed the Mean Field Networks, based on a feed-forward
network view of the mean field algorithm with fixed number of iterations. We
show that relaxing the restrictions on MFNs can improve inference efficiency
and discriminative performance. There are a lot of possible extensions around this
model and we are working on a few of them: (1) integrate learning graphical model
and learning inference model together; (2) relaxing the network
structure restrictions; (3) extend the method to other inference algorithms
like belief propagation.


\bibliography{paper}
\bibliographystyle{icml2014}

\end{document}